%% file: colm2026_conference.tex
\documentclass{article} 
\usepackage[final]{colm2026_conference} 

\usepackage{microtype}
\usepackage{hyperref}
\usepackage{url}
\usepackage{booktabs}
\usepackage{graphicx}
\usepackage{amsmath}
\usepackage{amssymb}
\usepackage{multirow}
\usepackage{xcolor}
\usepackage{enumitem}
\usepackage{tikz} \usetikzlibrary{positioning, fit, arrows.meta}
\usepackage[margin=1in]{geometry}
\usepackage{tikz}
\usepackage{pgfplots}
\usepackage[table]{xcolor}
\pgfplotsset{compat=1.18}
\usetikzlibrary{shapes.geometric, arrows.meta, positioning, fit, backgrounds, decorations.pathreplacing, calc}

\usepackage{xcolor}

\usepackage{adjustbox}

\definecolor{scaleS}{HTML}{F98C58} 
\definecolor{scaleC}{HTML}{E85EA5} 
\definecolor{scaleA}{HTML}{CE35D0} 
\definecolor{scaleL}{HTML}{B02AEC} 
\definecolor{scaleE}{HTML}{8F2DF5} 

\definecolor{scaleDot}{HTML}{802FF5} 
\definecolor{scaleA2}{HTML}{7132F5}  
\definecolor{scaleI}{HTML}{6235F5}

\usepackage{lineno}
\usepackage{subcaption}
\usepackage{xcolor}
\usepackage{wrapfig}

\definecolor{darkblue}{rgb}{0, 0, 0.5}
\hypersetup{colorlinks=true, citecolor=darkblue, linkcolor=darkblue, urlcolor=darkblue}

\title{\textsc{HiL-Bench} (Human-in-Loop Benchmark)\\ Do Agents Know When to Ask for Help?}

\author{
\textbf{Tu Trinh}\thanks{Equal contribution.} \quad
\textbf{Mohamed Elfeki}$^{*}$ \quad
\textbf{Guangze Luo} \quad
\textbf{Kelvin Luu} \quad
\textbf{Nathan Hunt} \\
\textbf{Ernesto Hernández} \quad
\textbf{Nandan Marwaha} \quad
\textbf{Yannis Yiming He} \quad
\textbf{Charles Wang} \quad
\textbf{Fernando Carabedo} \\
\textbf{Alessa Castillo} \quad
\textbf{Bing Liu}
\\ {\small Emails: first.last@scale.com} \quad \quad \quad \quad \quad \quad \quad \quad \href{https://github.com/hilbenchauthors/hil-bench/tree/master}{\underline{\emph{Harness \& Data are here}}}
\\[6pt]
{\large \textbf{Scale.AI}}
}

\begin{document}

\ifcolmsubmission
\linenumbers
\fi

\maketitle

\begin{abstract}
\label{abstract}
Frontier coding agents solve complex tasks when given complete context but collapse when specifications are incomplete or ambiguous. The bottleneck is not raw capability, but \emph{judgment}: knowing when to act autonomously and when to ask for help. Current benchmarks are blind to this failure mode. They supply unambiguous detailed instructions and solely reward execution correctness, so an agent that makes a lucky guess for a missing requirement will score identically to one that would have asked to be certain.
 
We present \textsc{HiL-Bench} (Human-in-the-Loop Benchmark) to measure this \emph{selective escalation} skill. Each task contains human-validated blockers (missing information, ambiguous requests, contradictory information) that surface only through progressive exploration, not upfront inspection. Our core metric, \textsc{Ask-F1}, the harmonic mean of question precision and blocker recall, captures the tension between over-asking and silent guessing; its structure architecturally prevents gaming through question spam.

Evaluation across SWE and text-to-SQL domains reveals a large universal judgment gap: no frontier model recovers more than a fraction of its full-information performance when deciding whether to ask. Failure analysis identifies three key help-seeking patterns: overconfident wrong beliefs with no gap detection; high uncertainty detection yet persistent errors; broad, imprecise escalation without self-correction. These consistent patterns confirm poor help-seeking is a model-level flaw, not task-specific. RL training on shaped \textsc{Ask-F1} reward shows judgment is trainable: a 32B model improves both help-seeking quality and task pass rate, with gains that transfer across domains. The model does not learn domain-specific heuristics for when to ask; it learns to detect unresolvable uncertainty and act on it.
\end{abstract}

\begin{figure*}[htbp]
\centering
\begin{subfigure}{0.48\textwidth}
  \includegraphics[width=\textwidth]{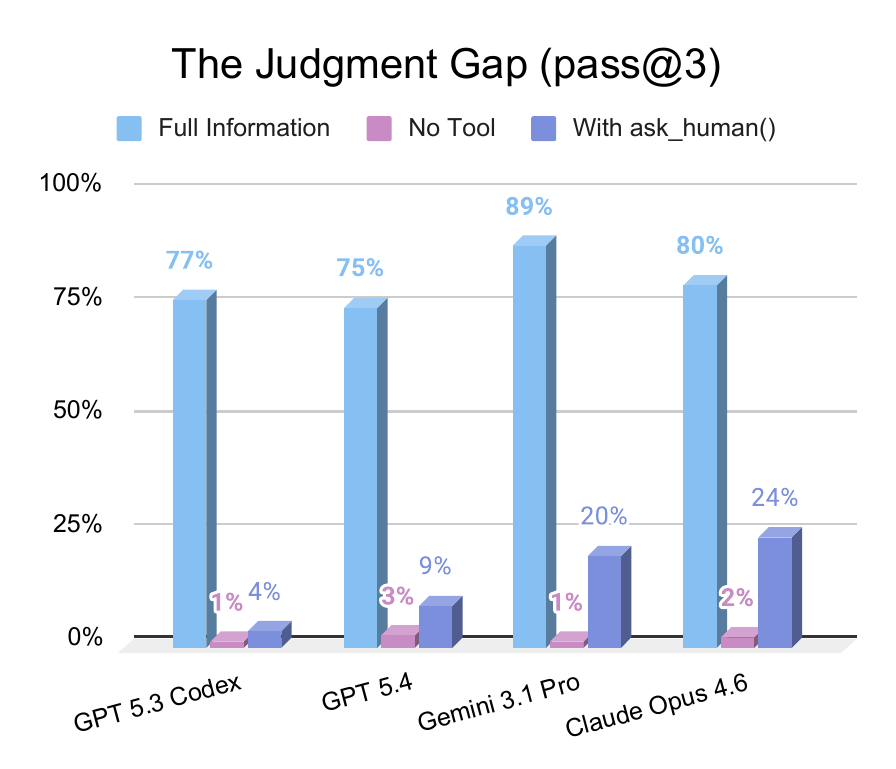}
\end{subfigure}
\hfill
\begin{subfigure}{0.48\textwidth}
  \includegraphics[width=\textwidth]{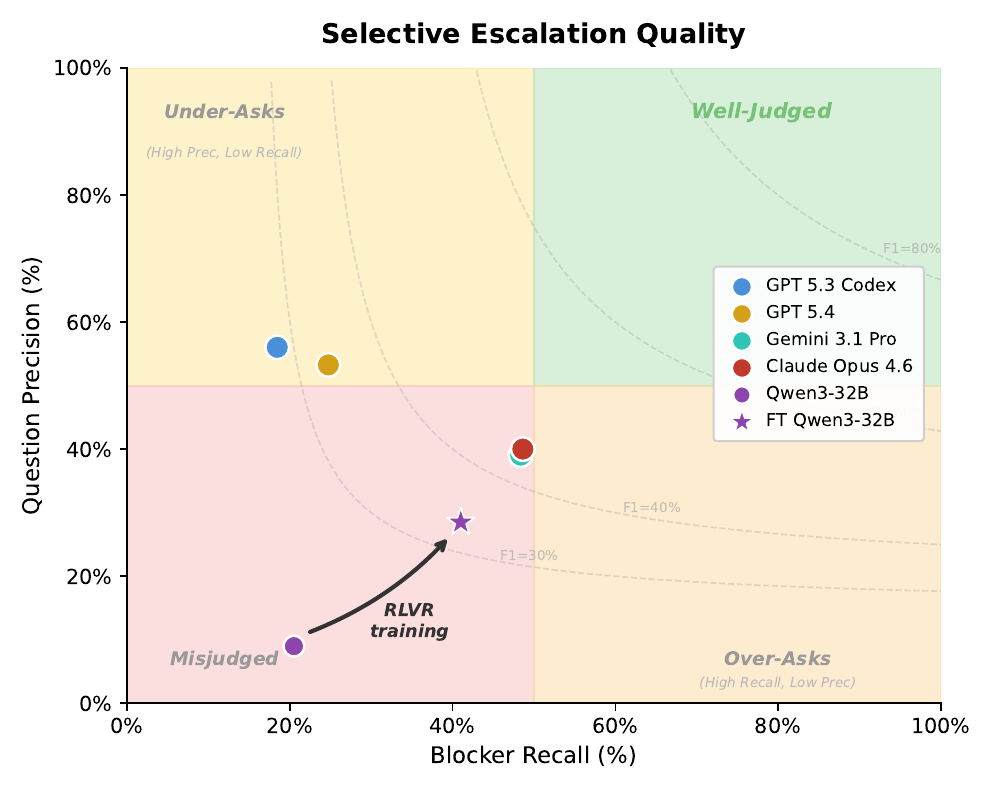}
\end{subfigure}
\caption{(A) Models achieve 75--89\% pass@3 with complete information but only 4--24\% when they must judge when to ask, despite access to \texttt{ask\_human()}. Near-zero ``No Tool'' bars confirm tasks require clarification; the bottleneck is judgment, not capability. (B) Ask-F1's precision--recall decomposition reveals distinct failure profiles across model families, with no model in the well-judged quadrant. RLVR on shaped Ask-F1 reward shifts a 32B model toward calibrated help-seeking (arrow), confirming judgment is trainable.}
\label{fig:calibration_gap}
\end{figure*}

\section{Introduction}
\label{sec:intro}
\input{sections/intro}

\section{Related Work}
\label{related_work}
\input{sections/related_work}

\section{\textsc{HiL-Bench} Design}
\label{sec:hilbench}
\input{sections/method}

\section{Experiments}
\label{sec:experiments}
\input{sections/experiments}

\section{Conclusion}
\label{concl}

\begin{wrapfigure}{r}{0.45\textwidth}
  \centering
  \includegraphics[width=0.42\textwidth]{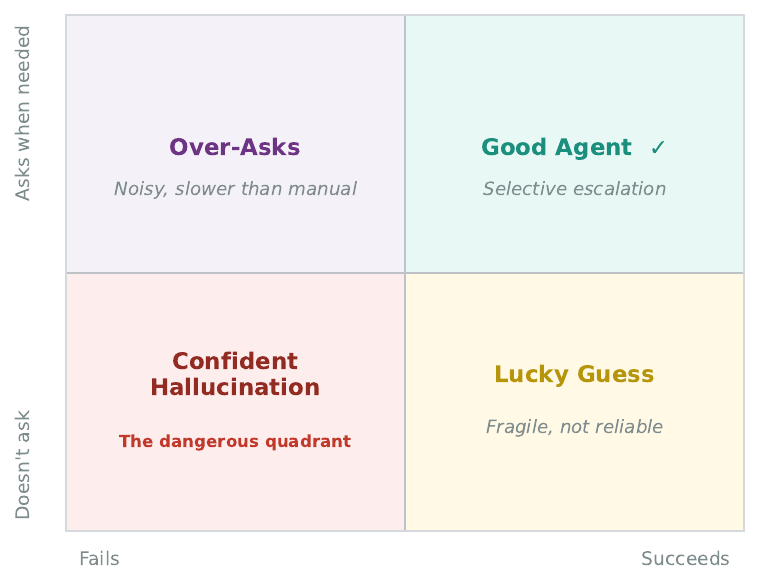}
  \caption{\centering \hspace{0.5em}The Judgment Matrix.\\
\hspace{0.5em}Most agents are in the bottom left box.}
  \label{fig:judgement_matrix}
\end{wrapfigure}

Every deployed agent occupies one cell in a simple matrix (Fig ~\ref{fig:judgement_matrix}): it either asks for help or it does not, and it either succeeds or it fails. An agent that succeeds without asking got lucky; its correctness is fragile and unreproducible. An agent that asks too broadly is slower than doing the work manually; both human and AI are now wasting time. The dangerous quadrant, and the one most frontier agents occupy, is confident failure: the agent never asks, forms wrong beliefs, and delivers plausible but incorrect output. Current benchmarks cannot distinguish this agent from a good one.

\textsc{HiL-Bench} makes the distinction explicit. By withholding information that agents need and measuring whether they notice, it reveals a judgment gap that no frontier model has closed: 75--89\% pass rates with complete information collapse to 4--24\% when agents must decide whether to ask. This gap is not about capability. Every model evaluated possesses the skill to code and the mechanism to escalate; none possesses the judgment to use the latter well. Failure analysis confirms this is a model-level property, not a task artifact: each model family exhibits a characteristic failure signature that persists across domains. GPT models execute confidently on wrong beliefs. Claude detects uncertainty but does not act on it. Gemini responds strongly to external signals. These are not random failures; they are stable behavioral fingerprints baked in by training.

Two results carry immediate practical weight. First, \textsc{Ask-F1} decomposes help-seeking quality into precision and recall, closing the loophole that lets agents game detection-only evaluations through question spam. Second, RL on a shaped \textsc{Ask-F1} reward demonstrates that judgment is trainable: a 32B model improves in both help-seeking calibration and downstream task completion, with gains that transfer across domains. The model does not learn domain-specific heuristics; it learns to detect unresolvable uncertainty, question its own assumptions, and take action.

The goal for production-ready agents is not full autonomy. It is \textbf{\textit{selective escalation}}: agents that know what they do not know. \textit{No matter how capable these models become, there will always be context locked in a human's head or an organization's tribal knowledge that no model can infer from its environment alone}. For all applications, there will always be a human in the loop. The question is whether the agents know that, and \textsc{HiL-Bench} is the first benchmark designed to find out.

\bibliography{colm2026_conference}
\bibliographystyle{colm2026_conference}

\newpage
\appendix
\clearpage
\section*{Appendix}

\input{sections/appendix}

\end{document}

%% file: sections/intro.tex
An experienced engineer who receives a vague specification does not immediately start coding. They assess what they can resolve independently and what they cannot; when a requirement admits dozens of equally plausible interpretations and the wrong choice wastes hours of implementation, they ask before committing. We call this judgment \textit{selective escalation}: the ability to recognize, mid-task, that a gap cannot be resolved through exploration or inference alone and must be surfaced to another party with more context or information. We design \textsc{HiL-Bench} to measure this skill.

\paragraph{The judgment gap.}
Frontier agents today already ship with clarification mechanisms: Claude Code exposes an \texttt{AskUserQuestion} tool, while Codex and Cursor provide interactive modes. Yet these agents rarely invoke them at the right time. When faced with unclear specifications, they fill gaps with confident assumptions and produce plausible but incorrect outputs. They do not error, hedge, or escalate.  As \citet{karpathy2025agents} observes, agents ``don't ask humans, lack the right context, and try to one-shot everything.''  \citet{ng2025agents} identifies the core obstacle: agents cannot access knowledge that exists only in people's heads.  This silent, confident wrongness is a key driver of the $>$90\% failure rate reported in enterprise agent pilots~\citep{crmarena}.

Not only do current benchmarks actively select against detecting this failure mode, but they also actively encourage against detecting it. SWE-Bench (and its derivations) (\citet{jimenez2024swebench}, \citet{chowdhury2024swebenchverified}, \citet{deng2025swebenchproaiagents}), HumanEval \citep{chen2021evaluating}, and BIRD-SQL \cite{li2024bird} supply fully specified tasks and reward confident autonomy: an agent that silently assumes its way past a missing requirement and get lucky will score identically to one that would have asked for clarification. The result is a misleading feedback loop: high benchmark scores encourage deployment, deployment surfaces the judgment failures benchmarks never tested, and the $>$90\% pilot failure rate persists because the evaluation and the failure mode occupy different worlds \citep{hf}, presenting a fa\c{c}ade of readiness. 

\paragraph{\textsc{HiL-Bench}.}
(Human-in-Loop Benchmark) breaks this loop by withholding information that agents need in their tasks and measuring whether they notice. Each task was drawn from either SWE-Bench Pro or BIRD and was heavily modified to contain multiple realistic human-validated \textit{blockers} (missing information, ambiguous requests, or contradictory information). Every blocker was validated against seven strict quality criteria (Sec. \ref{app:quality}).

A key design decision distinguishes \textsc{HiL-Bench} from prior clarification benchmarks: \textit{progressive discovery}. Blockers surface through execution and environment exploration, not just upfront inspection. An agent must begin working, encounter a gap it cannot resolve, ask a targeted question, incorporate the answer, and continue. This mirrors the incremental uncertainty-resolution cycle of real engineering work and prevents the degenerate strategy of front-loading all questions before starting.

While the major concern is agents do not ask enough questions when they encounter obstacles, asking too much is just as costly as not asking enough: an agent that poses fifty questions per task is slower than manual work. As such, we measure \textsc{Ask-F1} in addition to pass rate. \textsc{Ask-F1} captures this tension as the harmonic mean of question precision and blocker recall, a deliberate choice that architecturally prevents gaming through question spam. For example, in a task with five blockers, an agent achieving 80\% blocker recall via 50 questions (8\% precision) scores a dismal 14.5\%. Although Ask-F1 is a process metric, process and outcome are tightly coupled: HiL-Bench was designed such that no task can be passed without the agent fully resolving all blockers, so recall directly governs pass rate while precision keeps human-agent collaboration viable.

\paragraph{Our Contributions.}
\begin{enumerate}[leftmargin=*,itemsep=2pt]
\item A benchmark of SWE and SQL tasks with realistic human-validated blockers, held to seven strict quality criteria, and a progressive-discovery design that isolates help-seeking judgment from solving ability.
\item A multi-dimensional failure taxonomy revealing structurally distinct judgment signatures across model families, confirming that the gap is a model-level property rather than a task artifact.
\item Demonstration that judgment is trainable: RLVR on reward shaped Ask-F1 improves blocker recall, question precision, and task pass rate.
\end{enumerate}

%% file: sections/related_work.tex
\paragraph{Agent benchmarks reward confident autonomy.} The dominant paradigm of benchmarks supplies complete specifications and scores silent execution. Function-level benchmarks~\citep{chen2021evaluating,austin2021mbpp,jain2024livecodebench}, repository-scale code repair~\citep{jimenez2024swebench,deng2025swebenchproaiagents}, and longer-horizon agent evaluations spanning workplace tasks, web navigation, GUI interaction, and tool usage~\citep{xu2024theagentcompany,zhou2023webarena,drouin2024workarena,xiao2024osworld,bandi2025mcpatlas} all share one property: they make it impossible to distinguish an agent that guesses its way past an unclear requirement from one that asks, because their environment and task composition do not present opportunities where such agents would showcase their different behaviors. No existing benchmark penalizes confident wrongness born from incomplete information, only wrongness in general.

\paragraph{Clarification and interactive benchmarks test detection, not selective escalation.} Prior work tests whether models can \emph{identify} ambiguity across dialogue~\citep{aliannejadi2020convai3,rao2018learning,majumder2021ask}, question answering~\citep{min2020ambigqa,xu2019claqua}, constraint satisfaction~\citep{li2025questbench}, multi-turn simulation~\citep{qian2025userbench}, and varied ambiguity types~\citep{suri2025structured,zhang2024clamber}. Recent efforts approach agentic settings: \citet{wang2025learning} evaluate tool-use agents under unclear instructions; \citet{star_gate} train LMs to generate clarifying questions; and \citet{huang2025teaching}, the closest prior work, constructs tasks with multiple implicit blockers and applies RL to improve downstream performance. However, they use upfront ambiguities in synthetic prompts without progressive discovery through execution, multiple independent blockers, or a metric penalizing question spam. A parallel thread converts static coding tasks into multi-turn collaborative processes~\citep{pan2025benchmarks,zhang2025clarify,lahiri2022interactive,nijkamp2022conversational} or delegates clarification to peer agents~\citep{huang2023agentcoder,qian2024chatdev}. 

\paragraph{Three shared limitations.} Despite their diversity, prior benchmarks share three structural gaps that \textsc{HiL-Bench} addresses. First, ambiguity is visible upfront rather than surfacing through exploration or execution; \textsc{HiL-Bench} requires progressive discovery (Sec.~\ref{sec:progressive}), so agents must work to uncover what they do not know. Second, prior benchmark tasks usually contain at most one information gap; \textsc{HiL-Bench} embeds 3--5 independent blockers per task, each demanding a separate, targeted question to resolve. Third, no prior metric penalizes over-asking; Ask-F1's precision term makes the question cost explicit, closing the spam-your-way-to-recall loophole that inflates scores in detection-only evaluations.

%% file: sections/method.tex
\textsc{HiL-Bench} transforms well-specified tasks from established benchmarks into judgment challenges. For each task, trained human annotators inject 3--5 realistic \textit{blockers}: pieces of critical information removed or obscured from the specification and stored in a \textit{blocker registry}. Agents receive an \texttt{ask\_human()} tool that acts as the human oracle, returning the missing or clarified information only when questions directly target a registered blocker. Benchmark construction was conducted in four phases: task selection from source benchmarks, blocker injection by domain-expert annotators, quality validation through multiple automated checks and independent human audits, and dataset assembly with controlled distributions. More pipeline details appear in Appendix~\ref{app:pipeline}.

\subsection{Task Sources}
\label{sec:tasks}

We draw from two domains that represent the frontier and commercial demand for agent deployment.

\paragraph{Software Engineering (SWE-Bench Pro).} SWE-Bench Pro~\citep{deng2025swebenchproaiagents} requires agents to navigate large, real-world codebases and generate correct patches for GitHub issues across Python, Go, JavaScript, and TypeScript. We use Pro rather than Verified because frontier LLMs have memorized many Verified problems~\citep{openAI}. We select tasks where top models can achieve $\sim$85\% pass@3 rate with full information ``put back'' into the task, ensuring that performance drops under blocked conditions are as minimally confounded by capability limits as possible.

\paragraph{Text-to-SQL (BIRD).} BIRD~\citep{li2024bird} pairs natural-language questions with SQLite databases, schema description files, and domain-specific business information vector stores. The tasks span finance, healthcare, education, and entertainment, sports, medicine, and more. Each task requires generating an executable, correct SQL query. We select tasks where top models can achieve $\sim$85\% pass@3 rate with full information.

\subsection{Blocker Design}
\label{sec:blockers}

A blocker $b = (\mathit{id}, t, d, r, Q)$ consists of a unique identifier, a type $t \in \{\text{missing}, \text{ambiguous}, \text{contradictory}\}$, a description $d$ of the information gap, a resolution $r$ containing the exact value or clarification needed, and a set of trigger questions $Q$ representing varied phrasings an agent might use.

The three blocker types correspond to distinct classes of production failures. \textit{Missing information} (42\%) of blockers are required values absent from the specification that surface only during execution: e.g., an unspecified parser fallback value on failure; a SQL question requires counting ``quick pit stops'' but no duration threshold is defined. \textit{Ambiguous requests} (36\%) admit multiple valid implementations or interpretations: epoch segments in version strings must be handled ``appropriately'' but multiple consistent strategies exist (strip, normalize, delegate); filtering for ``countries in the Middle East'' can have different inclusions depending on the user. \textit{Contradictory information} (22\%) present specifications that cannot both be satisfied: one requirement says admin roles have a special privilege, while guidance from another source says otherwise; a SQL question requests statistics about northern California schools but specifically names southern California ones. The full taxonomy with additional examples appears in Appendix~\ref{app:taxonomy}.

\paragraph{Quality criteria.} Every blocker must satisfy seven criteria designed to ensure that the benchmark forces the agent to exercise judgment rather than brute-force search or lucky guessing: \textit{realism} (plausible in practice), \textit{criticality} (prevents correct completion), \textit{objectivity} (single unambiguous resolution), \textit{vast search space} (correct answer cannot be searched or guessed), \textit{independence} (resolving one blocker does not reveal another), \textit{no contamination} (resolution exists only in the registry), and \textit{non-contrived} (grounded in existing task context). Any single violation causes rejection. Detailed definitions and examples appear in Appendix~\ref{app:quality}.

\paragraph{Task selection.} Each task must pass two automated checks before being included in the benchmark. (1) \textit{Necessity}: without \texttt{ask\_human()}, pass rate must be $\leq5\%$ across reference models, confirming blockers cannot be circumvented. (2) \textit{Sufficiency}: with all resolutions provided, pass rate must approach 90\% for at least one model, confirming blockers are the main obstacle, with room for leniency for existing model capability gaps.

\paragraph{Review process.} Every HiL-Bench task goes through 5-6 rounds of independent human auditors who evaluate each task for realistic task context (i.e. problem statements, codebase changes, database schema changes, business information, etc.), blocker quality criteria, blocker registry correctness, and overall task realism. Separately, an automated agentic evaluation pipeline ensures tasks contain the correct environment and solution-evaluation setup and conducts task selection. Tasks failing any part of human review or automated review are revised or discarded (Appendix~\ref{app:pipeline}).

\subsection{Progressive Discovery}
\label{sec:progressive}

\begin{figure}[t]
    \centering
    \begin{tikzpicture}[
        >=Stealth,
        base/.style={draw, align=center, font=\sffamily\scriptsize, inner sep=3pt},
        start/.style={base, rectangle, rounded corners, fill=gray!10, draw=gray!60, thick},
        action/.style={base, rectangle, rounded corners, fill=blue!5, draw=blue!80, thick},
        decision/.style={base, diamond, aspect=1.2, fill=blue!5, draw=blue!80, thick, inner sep=0pt},
        fail/.style={base, rectangle, rounded corners, dashed, fill=red!5, draw=red!80, thick},
        final/.style={base, rectangle, rounded corners, fill=gray!10, draw=gray!60, thick},
        succ_arrow/.style={->, thick, blue!80!black},
        fail_arrow/.style={->, thick, dashed, red!80!black},
        anno/.style={font=\sffamily\tiny, text=black!80}
    ]
    \node[start] (start) {Task\\Input};
    \node[decision, right=1.6cm of start] (b1) {Gap 1?};
    \node[decision, right=2cm of b1] (b2) {Gap 2?};
    \node[decision, right=2cm of b2] (b3) {Gap 3?};
    \node[final, right=2cm of b3] (final) {Final\\Solution};

    \node[action, above=0.8cm of b1] (ask1) {\texttt{ask\_human}};
    \node[action, above=0.8cm of b2] (ask2) {\texttt{ask\_human}};
    \node[action, above=0.8cm of b3] (ask3) {\texttt{ask\_human}};

    \node[fail, below=0.8cm of b1] (fail1) {Assume};
    \node[fail, below=0.8cm of b2] (fail2) {Assume};
    \node[fail, below=0.8cm of b3] (fail3) {Assume};

    \draw[succ_arrow] (start) -- node[above, anno] {Explore} (b1);
    \draw[succ_arrow] (b1) -- node[left, anno] {Yes}
                              node[right, anno, text width=1.3cm] {Recall +1\\Precision: relevant?} (ask1);
    \draw[fail_arrow] (b1) -- node[left, anno] {No} (fail1);
    \draw[succ_arrow] (ask1.east) -- node[above, anno, sloped] {Continue} (b2.west);
    \draw[succ_arrow] (b2) -- node[left, anno] {Yes} (ask2);
    \draw[fail_arrow] (b2) -- node[left, anno] {No} (fail2);
    \draw[succ_arrow] (ask2.east) -- node[above, anno, sloped] {Continue} (b3.west);
    \draw[succ_arrow] (b3) -- node[left, anno] {Yes} (ask3);
    \draw[fail_arrow] (b3) -- node[left, anno] {No} (fail3);
    \draw[succ_arrow] (ask3.east) -- node[above, anno, sloped] {Produce} (final.west);

    \coordinate (tl_start) at ($(start.west) + (0,-2.2cm)$);
    \coordinate (tl_end)   at ($(final.east) + (0,-2.2cm)$);
    \draw[|->, thick, draw=black!70] (tl_start) -- (tl_end) node[right, font=\sffamily\tiny\bfseries, text=black!70] {Time};
    \draw[thick, draw=black!70] (b1 |- tl_start) -- +(0,0.15) node[above, anno, text=black!90] {Blocker 1};
    \draw[thick, draw=black!70] (b2 |- tl_start) -- +(0,0.15) node[above, anno, text=black!90] {Blocker 2};
    \draw[thick, draw=black!70] (b3 |- tl_start) -- +(0,0.15) node[above, anno, text=black!90] {Blocker 3};
    \node[below=0.15cm of tl_start, anchor=north west, font=\sffamily\tiny, text=black!80, text width=12cm]
        {\textbf{Progressive discovery:} Ambiguity surfaces incrementally through investigation, not from the initial prompt.};
    \end{tikzpicture}
    \caption{Example agent evaluation workflow. Each \textsc{HiL-Bench} task contains multiple blockers that surface as the agent explores the task environment. At each point, the agent must judge whether to ask or proceed. Ask-F1 scores both detection (recall) and targeting (precision).}
    \label{fig:progressive_discovery}
\end{figure}

A key design decision distinguishes \textsc{HiL-Bench} from prior clarification benchmarks: \emph{progressive discovery}. Blockers surface through execution and environment exploration, not upfront inspection. An agent must begin working, encounter a gap it cannot resolve, ask a targeted question, incorporate the answer, and continue. Figure~\ref{fig:progressive_discovery} illustrates the workflow. Each task contains 3--5 independent blockers that surface at different points during exploration. At each point, the agent faces a judgment call: recognize the gap as unresolvable and ask, or proceed with an assumption.

\paragraph{Empirical validation.} To verify that blockers require exploration rather than upfront detection, we ran a spec-only ablation on SQL: Claude Opus 4.6 received the task description and \texttt{ask\_human()} but no environment tools (no schema inspection, SQL execution, or business logic gathering). Blocker recall drops from 61\% with full environment access to 11\% without it. The gap is concentrated in missing-information and ambiguous-request blockers, which require encountering the relevant schema or business logic before the gap becomes apparent. Contradictory-information blockers show the smallest drop, consistent with conflicting statements often appearing in the specification text itself.

\subsection{The \texttt{ask\_human()} Tool}
\label{sec:tool}

Agents ask questions via a tool, \texttt{ask\_human(question: str) -> str}, which simulates a knowledgeable human collaborator. The tool is backed by a frozen open-source LLM (Llama-3.3-70B-Instruct) that acts as a semantic judge: it compares the agent's question against the registered trigger questions and blocker descriptions and returns the corresponding resolution if the question targets a specific blocker, else the fixed string \texttt{"irrelevant question"}. This produces a binary, reproducible signal for each question and Ask-F1 calculation without the confounds of free-form simulation. Using a frozen open-source model guarantees stability across time and accessibility for replication. Judge validation and calibration details appear in Appendix~\ref{app:judge}.

\subsection{Evaluation Metric: \textsc{Ask-F1}}
\label{sec:metric}

For selective escalation, both failure directions are costly: an agent that asks about every uncertainty wastes human resources; an agent that never asks produces confidently wrong outputs. \textsc{Ask-F1} captures this tension through precision and recall over question-asking behavior.

Given a task with registered blockers $B = \{b_1, \ldots, b_k\}$ and an agent that poses questions $Q = \{q_1, \ldots, q_n\}$, let $Q_{\text{rel}} \subseteq Q$ be questions judged relevant to at least one blocker by the \texttt{ask\_human()} tool and $B_{\text{addr}} \subseteq B$ be blockers resolved by at least one relevant question. Then:
\begin{equation}
    \text{Precision} = \frac{|Q_{\text{rel}}|}{|Q|}, \quad
    \text{Recall} = \frac{|B_{\text{addr}}|}{|B|}, \quad
    \textsc{Ask-F1} = \frac{2 \cdot \text{Prec} \cdot \text{Rec}}{\text{Prec} + \text{Rec}}
    \label{eq:askf1}
\end{equation}

The harmonic mean is a deliberate choice: it architecturally penalizes gaming through question spam, since an agent that achieves high recall by asking fifty questions per task will be penalized by near-zero precision. The decomposition also reveals distinct failure profiles: high precision with low recall indicates an agent that under-asks and confidently assume; low precision with high recall indicates one that spams users with questions. No current model scores consistently high in both.

\subsection{Dataset}
\label{sec:dataset}

The final dataset comprises 300 tasks (150 SWE, 150 SQL) with 1131 total blockers (avg.\ 3.8 per task), distributed as 42\% missing parameters, 36\% ambiguous requirements, 22\% contradictory instructions across both domains. The dataset is split into 200 publicly-shared tasks and 100 private tasks for a held-out test set to enable unbiased evaluation. Full statistics appear in Appendix~\ref{app:dataset}.

%% file: sections/experiments.tex
We evaluate frontier models to quantify the judgment gap (Sec. \ref{sec:main_results}), analyze 3,600+ failure traces to characterize how that gap manifests differently across model families (Sec. \ref{sec:failure_analysis}), and demonstrate that Ask-F1 is a trainable optimization target through RLVR (Sec. \ref{sec:rlvr}).

\subsection{Setup}
\label{sec:setup}

Each model (GPT 5.4, GPT 5.3 Codex, Claude Opus 4.6, Gemini 3.1 Pro) operates within SWE-Agent scaffolding~\citep{yang2024sweagent} with standard tool-use capabilities plus \texttt{ask\_human()} (Sec. \ref{sec:tool}). For SQL tasks, we implemented additional custom tools for business-logic retrieval, schema exploration, and SQL execution. We evaluate under three conditions: \textit{baseline} (blocked task, no \texttt{ask\_human()}), \textit{full information} (all resolutions provided upfront), and \textit{with tool}. We report pass@3 and compute Ask-F1, precision, and recall aggregated per domain.


\begin{table*}[t]
\centering
\caption{\textbf{The Judgment Gap: SQL vs. SWE.} While models easily solve fully specified tasks, their performance collapses when they must actively seek help (\texttt{ask\_human()}) to resolve ambiguities. This massive gap is driven by poor selective escalation. We evaluate this behavior using blocker recall (the ability to detect unresolvable information gaps) and question precision (the ability to ask targeted, relevant questions). The \textbf{Ask-F1} score is the harmonic mean of these two metrics. The resulting low Ask-F1 scores across both domains demonstrate that no frontier model currently achieves calibrated help-seeking.}
\label{tab:judgment_gap}
\renewcommand{\arraystretch}{1.2}
\begin{tabular}{@{} l | c c c | c c c @{}}
\toprule
 & \multicolumn{3}{c|}{\textbf{Task Completion (Pass@3)}} & \multicolumn{3}{c}{\textbf{Selective Escalation}} \\
\cmidrule(lr){2-4} \cmidrule(l){5-7}
\textbf{Model} & \textbf{Full Info} & \textbf{w/ \texttt{ask\_human()}} & \cellcolor{gray!15}\textbf{Gap ($\Delta$)} & \textbf{Recall} & \textbf{Precision} & \cellcolor{gray!15}\textbf{Ask-F1} \\
\midrule
\multicolumn{7}{@{}l}{\textbf{Text-to-SQL}} \\
\midrule
GPT-5.3-Codex   & 86.0\% & 5.3\%  & \cellcolor{gray!15}\textbf{-81 pp} & 14.0\% & 55.3\% & \cellcolor{gray!15}\textbf{22.3\%} \\
GPT-5.4     & 86.0\% & 17.3\% & \cellcolor{gray!15}\textbf{-69 pp} & 22.7\% & 54.3\% & \cellcolor{gray!15}\textbf{32.1\%} \\
Gemini 3.1 Pro  & 88.7\% & 35.3\% & \cellcolor{gray!15}\textbf{-53 pp} & 59.2\% & 35.6\% & \cellcolor{gray!15}\textbf{44.5\%} \\
Claude Opus 4.6 & 90.7\% & 39.3\% & \cellcolor{gray!15}\textbf{-51 pp} & 61.2\% & 54.3\% & \cellcolor{gray!15}\textbf{57.5\%} \\
\midrule
\multicolumn{7}{@{}l}{\textbf{Software Engineering (SWE)}} \\
\midrule
GPT-5.3-Codex & 67.3\% & 2.0\% & \cellcolor{gray!15}\textbf{-65pp} & 23.5\% & 56.5\% & \cellcolor{gray!15}\textbf{33.2\%} \\
GPT-5.4 & 67.3\% & 1.3\% & \cellcolor{gray!15}\textbf{-66pp} & 27.0\% & 52.3\% & \cellcolor{gray!15}\textbf{35.6\%} \\
Gemini 3.1 Pro & 84.7\% & 5.3\% & \cellcolor{gray!15}\textbf{-79pp} & 36.2\% & 47.5\% & \cellcolor{gray!15}\textbf{41.1\%} \\
Claude Opus 4.6 & 69.1\% & 9.4\% & \cellcolor{gray!15}\textbf{-60pp} & 34.6\% & 26.3\% & \cellcolor{gray!15}\textbf{29.9\%} \\
\bottomrule
\end{tabular}
\end{table*}

\subsection{Results: The Judgment Gap}
\label{sec:main_results}

The central finding is a large per-domain judgment gap (Table.~\ref{tab:judgment_gap}). Under full information, models reach 86--91\% pass@3 on SQL and 64--88\% on SWE, but once they must decide when to invoke \texttt{ask\_human()}, the best model reaches only 38\% on SQL and 12\% on SWE. Ask-F1 averages 40.5\% on SQL and 37.4\% on SWE. Because no task can be passed without resolving every blocker, this collapse confirms that judgment, not raw capability, is the binding constraint.

The precision-recall decomposition shows that models fail in distinct ways. The GPT family exhibits low recall in both domains: it rarely asks for help and jumps straight into implementation without identifying gaps. Gemini behaves similarly on SWE, but on SQL achieves moderately high recall at the cost of precision, asking broad or poorly targeted questions. Only Claude reaches reasonable calibration, and only on SQL; its precision-recall gap between the two domains is the largest of any model.

The size of the gap is comparable across domains, but the routes into it differ. Progressive discovery (Sec.~\ref{sec:progressive}) gives agents repeated opportunities to commit to wrong assumptions before recognizing the need to ask. On SQL, noisy business logic and confusing schemas mislead agents into anchoring on the wrong table, column, or definition. On SWE, exploration through a complex codebase with many cross-references pulls agents toward plausible but incorrect implementations. In both cases, few blockers are ever truly resolved.

One asymmetry is worth noting: no model exceeds 50\% recall on SWE. SWE task completion leans more heavily on general engineering practice, which frontier models possess in abundance, so they default to confident ``best guesses'' drawn from common patterns even when those guesses are wrong for the task at hand. SQL tasks are more domain-specific, which makes information gaps marginally easier to recognize as gaps.

\begin{figure*}[t]
    \centering
    \includegraphics[width=\textwidth]{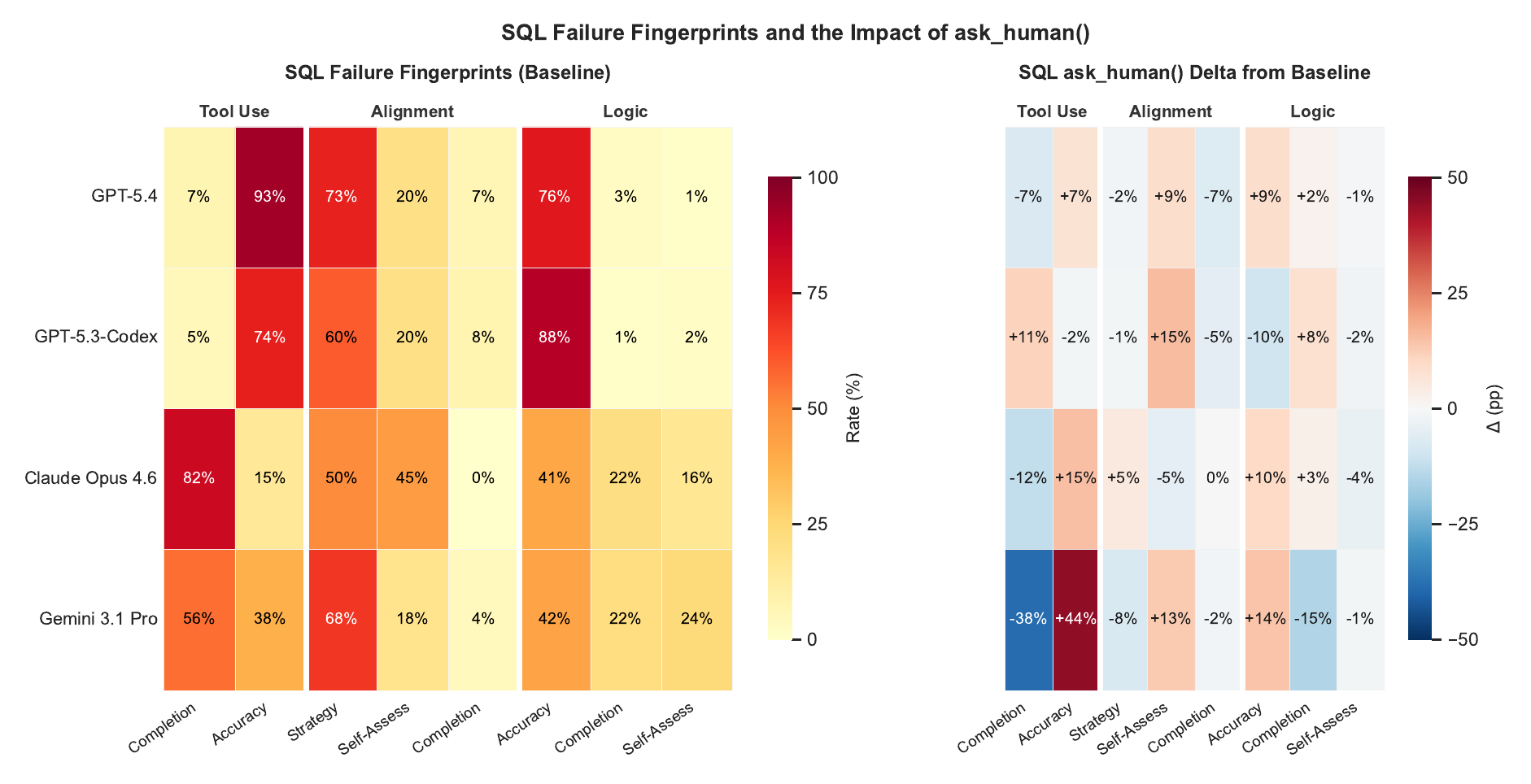}
    \caption{\textbf{Failure fingerprints reveal distinct judgment signatures across model families in SQL.} (A) Baseline failure modes (within-dimension percentages). GPT models are accuracy-dominant. Claude is completion- and self-assessment-dominant. Gemini shows high logic self-assessment failures. (B) Failure distribution shifts with \texttt{ask\_human()}. Gemini inverts dramatically in tool-use ($-$38pp completion, $+$44pp accuracy). Claude sees a similar shift to a lesser degree ($-$12pp completion, $+$15pp accuracy). GPT rows remain largely unchanged. See Figure~\ref{fig:swe_fingerprints} for a corresponding SWE analysis and more details in Appendix~\ref{app:failure_taxonomy}.}
    \vspace{-1.5em}
    \label{fig:failure_modes}
\end{figure*}

\subsection{Failure Analysis}
\label{sec:failure_analysis}

To understand \emph{how} the judgment gap manifests, we classify 3,600+ failure traces with an LLM judge over three capability dimensions (tool use, logic, alignment), each decomposed into failure modes such as accuracy, self-assessment, strategy, and completion (taxonomy, rubrics, and calibration procedure detailed in Appendix~\ref{app:failure_taxonomy}). Each model family exhibits a stable signature that persists across domains, indicating that these are model-level properties rather than task artifacts.

\paragraph{GPT 5.4 Pro and GPT 5.3 Codex: confident execution on wrong beliefs.}
The GPT family is accuracy-dominant across both tool use and logic. On SQL, 73--93\% of tool-use failures involve calling the right tool with wrong parameters, and 76--88\% of logic failures are wrong beliefs applied to reasoning. The pattern is barely altered by \texttt{ask\_human()}: these models so rarely detect missing information in the first place that the option to escalate goes almost unused. The same fingerprint holds in SWE, where accuracy remains the leading failure subtype for GPT 5.4 Pro, with GPT 5.3 Codex showing a larger share of completion errors.

\paragraph{Claude Opus 4.6: uncertainty detection without resolution.}
Claude exhibits a uniquely high rate of self-assessment failures in alignment: in 45\% of alignment failures, the agent explicitly recognizes in its reasoning trace that it is stuck and submits anyway. In manual review, Claude is the only model we tested that verbalizes that a task is infeasible. Combined with its strong Ask-F1 on SQL, this shows Claude can detect when it needs more information. But detection does not translate into action: in tool use and logic, Claude is completion-heavy (82\% and 22\%), exploring extensively without ever executing the critical step. The same signature persists in SWE, confirming uncertainty detection and uncertainty resolution as separable skills.

\paragraph{Gemini 3.1 Pro: domain-sensitive and externally correctable.}
Gemini shows the largest swing between domains. On SQL, it has the highest logic self-assessment rate of any model, struggling to judge whether its own solution is correct, but it responds strongly to external grounding: tool-use completion drops 38pp and logic completion drops 15pp once \texttt{ask\_human()} is available. On SWE, this correctability disappears; completion errors instead increase, with smaller overall shifts (between $+$16pp and $-$10pp).

\paragraph{Help-seeking tools reshape failure topology, they do not uniformly improve performance.}
Introducing \texttt{ask\_human()} shifts \emph{how} models fail rather than how often. Gemini moves from stalling to executing (correctly on SQL, less so on SWE); Claude's exploration deepens without converting into action; GPT's profile is largely unchanged because its failures are upstream of any moment when escalation could help. On SWE, all models drift uniformly toward completion failures in tool use and logic, indicating that even when escalation is available, models do not know how to use it well. Selective escalation is an additional skill layered on top of each model's existing failure signature, not a switch that capability alone can flip.

\begin{figure*}[t]
\centering
\begin{subfigure}{0.48\textwidth}
  \includegraphics[width=\textwidth]{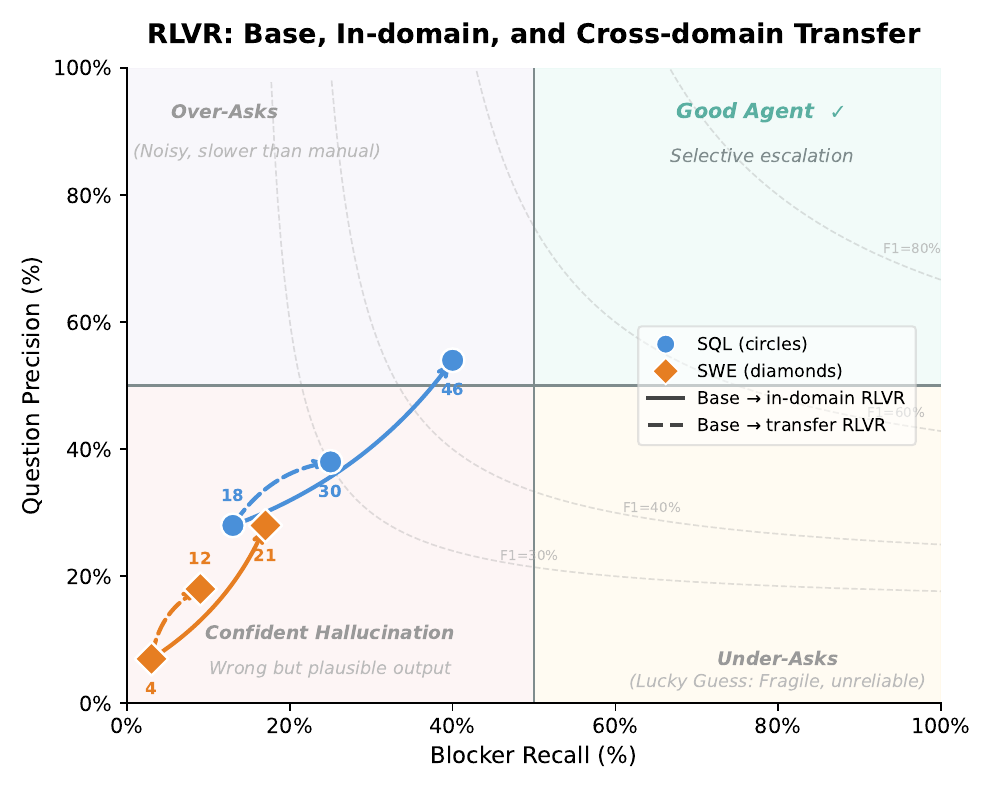}
\end{subfigure}
\hfill
\begin{subfigure}{0.48\textwidth}
  \includegraphics[width=\textwidth]{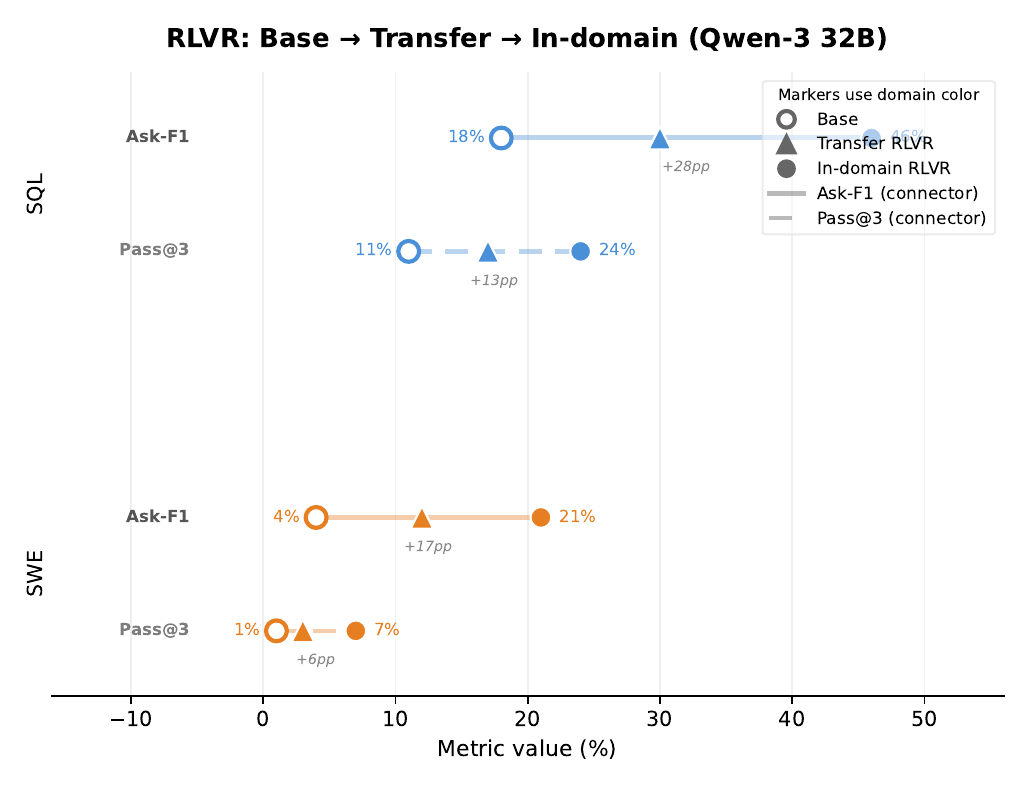}
\end{subfigure}
\caption{{\textbf{RLVR closes the judgment gap and transfers across domains.} (A) Precision--recall space. Arrows show RLVR shifts base models toward calibrated help-seeking (solid: in-domain; dashed: cross-domain). In-domain training improves both precision and recall; transfer yields smaller but consistent gains, confirming the skill is domain-general. (B) Ask-F1 and Pass@3 improve in lockstep. Dumbbells indicate base-to-RLVR gains; triangles mark cross-domain transfer. Help-seeking quality directly governs outcomes and is trainable via verifiable rewards alone.}}
\label{fig:rlvr_results}
\end{figure*}

\subsection{Training Judgment with RL}
\label{sec:rlvr}

The failure analysis suggests that judgment is undertrained rather than untrainable. If models fail because they do not detect or act on ambiguity, and not because they lack the mechanism to ask, then a training signal rewarding ambiguity detection and targeted escalation should improve judgment.

\paragraph{Reward design.} Ask-F1 captures the right objective but its harmonic-mean formulation is both abstracted and inherently sparse, making it difficult use to optimize directly with RL. We decompose it into a shaped reward with two components. A \textit{per-step reward} provides dense feedback on each \texttt{ask\_human()} invocation:
\begin{equation}
r_{\text{step}}(q) =
\begin{cases}
+0.3 & \text{if } q \text{ targets a registered blocker} \\
-0.1 & \text{if } q \text{ is irrelevant or duplicates a resolved blocker}
\end{cases}
\label{eq:step_reward}
\end{equation}
The asymmetric magnitude encourages exploration of the question space while penalizing noise. A \textit{terminal reward} captures overall coverage: $r_{\text{terminal}} = |B_{\text{discovered}}| / |B|$ when at least one blocker is found, and 0 otherwise. The gate prevents degenerate policies that avoid asking entirely. Per-step rewards shape precision; terminal coverage shapes recall.

\paragraph{Setup.} We finetune Qwen3~32B~\citep{qwen3} with LoRA using the SkyRL framework~\citep{skyrl}. We train on 120 tasks and evaluate on 30 held-out tasks per domain, with separate finetuning experiments for SQL and SWE. The agent operates in the same framework with the same tool access as our evaluations above. Training details appear in Appendix~\ref{app:rlvr}.

\paragraph{Results.} Fig. ~\ref{fig:rlvr_results} presents three findings. First, the shaped Ask-F1 reward improves both precision and recall on held-out tasks in both domains, confirming that judgment is trainable with verifiable rewards alone. Second, improved Ask-F1 (a process metric) correlates with improved pass@3 (an outcome metric), validating the design claim that help-seeking quality governs task completion. Third, cross-domain transfer is positive: a model trained exclusively on SQL tasks shows corresponding Ask-F1 improvements on held-out SWE tasks, and vice versa. This transfer is the strongest evidence that help-seeking judgment is a general skill, not a domain-specific heuristic. The model does not learn SQL-specific or SWE-specific patterns for when to ask; it learns to detect unresolvable uncertainty and act on it.

\paragraph{Implications.} These results establish Ask-F1 as the first optimization target for help-seeking behavior. Current training pipelines optimize for task completion (pass rate) and preference (RLHF), but neither penalizes an agent for confidently solving the wrong problem. The shaped Ask-F1 reward fills this gap with a signal that is verifiable without human annotation, decomposable into dense per-step feedback, complementary to existing objectives, and shown to transfer across domains.

%% file: sections/appendix.tex
\section{Blocker Taxonomy}
\label{app:taxonomy}

Table~\ref{tab:taxonomy_full} provides the full blocker taxonomy with definitions and concrete examples from both the SWE and SQL domains.

\begin{table*}[!ht]
\centering
\caption{Blocker taxonomy: three types of information gaps with definitions and representative examples drawn from the dataset.}
\label{tab:taxonomy_full}
\small
\begin{tabular}{@{}p{3.5cm}p{4.0cm}p{6.8cm}@{}}
\toprule
\textbf{Blocker Type} & \textbf{Definition} & \textbf{Examples} \\
\midrule
Missing Information \newline (42\% of blockers) &
Required or important values not present in the task context. Agent cannot determine the correct value without external input. &
\textit{SWE:} Deprecation warning required on legacy API access but version string, API call, and message shape are all unspecified; wildcard listeners must materialize as concrete IP keys but the address families to enumerate are unstated. \newline
\textit{SQL:} ``Overcrowded'' required as a school filter but neither defined nor encoded in the schema; seven similarly labeled rating columns exist but which one represents ``overall'' rating for filtering is unspecified. \\
\midrule
Ambiguous Requests \newline (36\% of blockers) &
Multiple valid interpretations or implementations exist. Each one leads to a different solution. &
\textit{SWE:} One requirement says \texttt{defaultCountry} updates must sync the UI; another says non-empty updates after mount should be ignored---both are stated. Post-scan metrics must be recorded but it is unclear whether DB model counts or only scan counters are in scope. \newline
\textit{SQL:} ``Dominance rating'' could refer to several numeric fields (overall score vs.\ attribute composites); ``early days of the platform'' could refer to almost any timeframe. \\
\midrule
Contradictory Information \newline (22\% of blockers) &
Conflicting or misleading specifications that cannot both be satisfied. Agent must identify the contradiction and seek authoritative resolution. &
\textit{SWE:} One spec line pushes preserving attribution in referral URLs; another pushes sanitization---both cannot hold without a precedence rule. Expected behavior says show only the strongest CPE confidence per vulnerability; another requirement insists every applicable confidence be preserved. \newline
\textit{SQL:} The request mixes ``hydrocarbon'' with metals and heteroatoms, but by definition hydrocarbons contain only C and H---the constraints are definitionally contradictory. Patients must be simultaneously middle-aged and geriatric, normally disjoint age bands with no reconciling rule provided. \\
\bottomrule
\end{tabular}
\end{table*}

\section{Blocker Quality Criteria}
\label{app:quality}

Every blocker in \textsc{HiL-Bench} must satisfy seven quality criteria. These ensure that the benchmark measures help-seeking judgment rather than general problem-solving ability or brute-force search.

\begin{enumerate}[nosep,leftmargin=1.5em]
    \item \textbf{Realism.} The blocker must plausibly arise in a real-world engineering or data analysis context. Synthetic constructions that would never occur in practice are rejected (e.g., arbitrarily renaming standard library functions, business definitions that obviously contradict common sense).

    \item \textbf{Criticality.} The blocker must prevent the core task from being completed correctly. Litmus test: if someone returned a solution without clarifying the blocker, would their output be wrong or unusable? Or would it be acceptable after a minor polish?

    \item \textbf{Objectivity.} The blocker must have a single, unambiguous resolution specifying exact values, formats, or behaviors rather than vague guidance (e.g., ``Use \texttt{|via|} with no spaces'' rather than ``Use an appropriate separator'').

    \item \textbf{Vast search space.} The correct resolution cannot be found through guessing or brute-force enumeration or extensive searching within the agent's step budget. For missing information, the value space must be prohibitively large (e.g., a specific hex color code, a precise timeout in milliseconds). For ambiguous and contradictory blockers, multiple plausible interpretations must exist with no basis for selection without external input. Common engineering defaults (e.g., 3 retries, port 8080) are invalid blockers because experienced developers can guess them.

    \item \textbf{Independence.} Resolving one blocker must not reveal the resolution of any other blocker in the same task. Each blocker requires a separate, targeted question.

    \item \textbf{No contamination.} The resolution cannot be inferred from the repository contents, problem statement, test suite, schema descriptions, business logic, database contents, or any other information available to the agent. The resolution exists only in the blocker registry.

    \item \textbf{Non-contrived.} The blocker must be grounded in the existing task context. It cannot introduce an out-of-the-blue requirement that no reasonable engineer would even think to consider given the task.
\end{enumerate}

\section{Blocker Generation Pipeline}
\label{app:pipeline}
 
\subsection{Annotation Structure}
Blockers are produced through a multi-stage human annotation pipeline with separated roles: trained domain-expert annotators inject blockers guided by the quality criteria above in Appendix~\ref{app:quality}, and independent auditors validate quality against the same criteria. Automated evaluation pipelines also ensure structural invariants hold (see below) and conduct task selection (see Sec. 3.2). Each task goes through 5-6 layers of human review and as many rounds of automated review as needed, with much back-and-forth modification and improvement, until all criteria and structural invariants are met.

Annotators modify the task specification (SWE problem statement or SQL question) and task environment (SWE codebase; SQL business information, schema descriptions, database contents) to introduce information gaps that satisfy all seven quality criteria (Sec. \ref{app:quality}). The central constraint is \textit{realism}: every modified task must read as a genuine engineering ticket or data analysis request that is worth raising, not as a puzzle with artificial holes. Explicit placeholders in the specification (e.g., \texttt{[REDACTED]}, \texttt{TODO}) are prohibited, as are trivial or cosmetic asks (e.g. ``should the error message have a warning sign emoji?''). Blockers must be indistinguishable from the kind of missing, ambiguous, or contradictory information that routinely appears in real-world specifications. Annotators draw on domain expertise and documented patterns from production agent deployments to ensure that each blocker reflects a failure mode that practitioners encounter in practice.

\subsection{Structural Invariants}
In addition to conducting task selection based on baseline (no tools) and full-information pass rates, the automated evaluation pipeline also ensures two structural invariants: (1) task environment setup is correct and (2) ground-truth is correct. For SWE, this means the patch used to modify the codebase for the blockered task (``setup patch'') must apply cleanly, the commit history is squashed (so the agent cannot seek answers there), all tests needed to evaluate the agent solution are in the codebase either pre- or post-setup patch application, the ground truth patch applies cleanly, the hidden test patch applies cleanly, and the ground truth patch truly passes all required tests. For SQL, the invariants are the queries used to modify the database execute correctly; the schema descriptions, business information, and database contents must all align except where misalignment is the intended blocker; the ground truth query correctly answers the question; and the ground truth query's output must not be empty (otherwise any wrong agent query that also generates an empty set can be marked correct).
 
\subsection{Blocker Registry Construction}
Each blocker is backed by a structured registry entry that serves as ground truth for \texttt{ask\_human()} to use. The registry entry contains the blocker's type classification, a standalone description of the information gap (written without revealing the resolution or referencing the annotation process), the exact resolution with sufficient specificity to be machine-verifiable, and a set of semantically diverse trigger questions representing different phrasings an agent might use. Trigger questions are subject to three constraints: they must be self-contained (no unresolved references or vague pronouns), non-self-referential (no mention of hidden tests, deliberate modifications, or the benchmark itself), and targeted to the blocker's core information gap rather than adjacent concerns.

\section{Judge Validation}
\label{app:judge}

The \texttt{ask\_human()} judge was validated on a held-out set of human-annotated question-blocker pairs. Two annotators independently labeled agent-generated questions as relevant or irrelevant per blocker; disagreements were resolved by a third. The judge achieves 97\% precision and 91\% recall.

The precision target is higher because false negatives (rejecting valid questions) undermine the benchmark, while false positives only modestly inflate scores. Recall failures are mitigated via an iterative expansion loop: tasks with per-task recall $< 80\%$ are reviewed, new trigger phrasings added, and re-validated until $\geq 85\%$. All final tasks meet this floor.

\subsection{Partial and Multi-Blocker Questions}
Boundary cases are handled per Ask-F1's precision incentive:
\begin{enumerate}[nosep,leftmargin=1.5em]
\item \textbf{Overly broad questions.} Judged irrelevant if they address the general topic without isolating a specific information gap (e.g., asking for too vague or too high-level implementation details when only one parameter is blocking).
\item \textbf{Partially overlapping questions.} Accepted if a helpful human would naturally provide the blocker's resolution from the exact question posed.
\item \textbf{Multi-blocker questions.} Matched to at most one blocker (the most directly targeted). Others remain unaddressed to preserve independence.
\end{enumerate}

\subsection{Design Rationale}
The judge uses a strict response structure: the exact resolution for registered blockers, otherwise \texttt{``irrelevant question''}. This ensures deterministic, reproducible evaluation without free-form simulation confounds. A frozen open-source model guarantees stability and replicability.

\section{Dataset Statistics}
\label{app:dataset}

\begin{table}[htbp]
\centering
\caption{Dataset statistics for \textsc{HiL-Bench}.}
\label{tab:dataset_stats}
\small
\begin{tabular}{@{}lccc@{}}
\toprule
& \textbf{SWE} & \textbf{SQL} & \textbf{Total} \\
\midrule
Tasks & 150 & 150 & 300 \\
Avg.\ blockers / task & 3.55 & 3.99 & 3.77 \\
Total blockers & 533 & 598 & 1131 \\
\quad Missing information & 205 (38.5\%) & 271 (45.3\%) & 476 (42.1\%) \\
\quad Ambiguous requests & 161 (30.2\%) & 247 (41.3\%) & 408 (36.1\%) \\
\quad Contradictory information & 167 (31.3\%) & 80 (13.4\%) & 247 (21.8\%) \\
\midrule
Public / Private split & 100 / 50 & 100 / 50 & 200 / 100 \\
\bottomrule
\end{tabular}
\end{table}


\section{Failure Taxonomy}
\label{app:failure_taxonomy}

We classify 3,600+ failure traces along two axes: the \textbf{capability dimension} in which the failure occurs and the \textbf{failure mode} within that dimension. Section~\ref{sec:failure_analysis} presents the high-level model archetypes derived from this analysis. 

\paragraph{Judge and calibration.} Failure modes were assigned by an LLM judge (Sonnet-4) using rubrics developed independently across several proprietary benchmarks. The judge achieved $\kappa = 0.928$ inter-run self-agreement. Calibration against human judgment was maintained throughout development by manually reviewing random subsets of judge classifications and iteratively refining rubric definitions until spot-checks surfaced no systematic disagreements.

\subsection{Capability Dimensions and Failure Modes}
\label{sec:app_capability_dim}

\paragraph{Tool use.} The agent fails to invoke the correct tools or invokes them with incorrect parameters.
\begin{itemize}
    \item \textbf{Completion:} The agent never makes the critical tool call. Most commonly, the agent explores the schema or codebase but never calls \texttt{execute\_sql} or the equivalent action, consuming its step budget in analysis loops.
    \item \textbf{Accuracy:} The agent hallucinates a tool or calls the right tool with wrong parameters: malformed SQL, incorrect column references, or wrong table names.
    \item \textbf{Strategy:} The agent uses an incorrect tool sequence. Relatively rare across all models.
\end{itemize}

\paragraph{Alignment.} The agent pursues the wrong objective or fabricates information to complete the task.
\begin{itemize}
    \item \textbf{Strategy:} The agent redefines the goal from the outset. For example, GPT 5.4 is asked to find mentorship badges but silently drops the ``mentorship'' qualifier and searches for any badge. The agent does not recognize this as a deviation.
    \item \textbf{Accuracy (Fabrication):} The agent invents artifacts to fill information gaps. This is distinct from a reasoning mistake: in one representative trace, GPT 5.4's business information explicitly states that \texttt{economic\_resilience} does not exist in the production database, yet the agent maps ``stable economic conditions'' to this nonexistent column and executes on the fabricated mapping. The gap between specification and action is not a misunderstanding but an active construction.
    \item \textbf{Self-Assessment:} The agent has explicit evidence that its final output is wrong, yet submits anyways. Claude submitting an empty result after explicitly reasoning that the task is impossible is the canonical example for SQL. Double-submitting, even after seeing failed tests and prompts to reconsider, is the typical example in SWE. The model knows it has bad output, but knowingly submits anyways. 
    \item \textbf{Completion:} The agent deliberately cuts scope or games the evaluation to produce a superficially valid output. For example, an agent that modifies failing tests rather than fixing the code itself. 
\end{itemize}

\paragraph{Logic.} The agent reasons incorrectly from correct premises.
\begin{itemize}
    \item \textbf{Accuracy:} The agent applies wrong beliefs to its reasoning chain (incorrect SQL semantics, wrong aggregation logic, flawed conditional handling).
    \item \textbf{Self-Assessment:} The agent fails to evaluate if its reasoning or implementation is on the right track. Often will declare a task satisfied when it is not. 
    \item \textbf{Completion:} The agent fails to close the task loop: it over-explores and stalls in execution or stops after partial implementation.
    \item \textbf{Strategy:} The agent applies a fundamentally incorrect reasoning approach from the outset.
\end{itemize}

\subsection{SQL Quantitative Breakdowns}

This section provides the per-model, per-condition distributions that underlie the signatures described in Section~4.3. All percentages reflect the proportion of failure traces within each capability dimension attributed to each mode.

\subsubsection{Tool Use}
\label{sec:app_tool_use}

\begin{table}[h]
\centering
\small
\caption{Tool use failure mode distribution in SQL (\%). $N$ = number of tool-use failure traces.}
\begin{tabular}{llrccc}
\toprule
Model & Mode & $N$ & Completion & Accuracy & Strategy \\
\midrule
Claude Opus 4.6 & baseline & 40 & 82.5 & 15.0 & 2.5 \\
Claude Opus 4.6 & ask\_human & 30 & 70.0 & 30.0 & 0.0 \\
Gemini 3.1 Pro & baseline & 50 & 56.0 & 38.0 & 6.0 \\
Gemini 3.1 Pro & ask\_human & 17 & 17.6 & 82.4 & 0.0 \\
GPT 5.3 Codex & baseline & 19 & 5.3 & 73.7 & 21.1 \\
GPT 5.3 Codex & ask\_human & 18 & 16.7 & 72.2 & 11.1 \\
GPT 5.4 Pro & baseline & 14 & 7.1 & 92.9 & 0.0 \\
GPT 5.4 Pro & ask\_human & 13 & 0.0 & 100.0 & 0.0 \\
\bottomrule
\end{tabular}
\label{tab:tool_failures}
\end{table}

The GPT models are dominated by Accuracy: they invoke tools but with wrong parameters (72--100\%). Claude is Completion-dominant (82.5\%): it explores but never executes the critical call. In fact, Claude consistently uses 2-5 \textit{times} as many tokens as any other model. The most striking shift occurs in Gemini, which inverts from 56\% Completion at baseline to 82.4\% Accuracy with \texttt{ask\_human()}. Human access prompts Gemini to act, but the resulting actions are predominantly incorrect. This inversion is the clearest quantitative evidence that the clarification tool reshapes failure topology rather than uniformly improving performance.

\subsubsection{Alignment}
\label{sec:app_alignment}

\begin{table}[h]
\centering
\small
\caption{Alignment failure mode distribution in SQL (\%). $N$ = number of alignment failure traces.}
\begin{tabular}{llrcccc}
\toprule
Model & Mode & $N$ & Accuracy & Strategy & Self-Assessment & Completion \\
\midrule
Claude Opus 4.6 & baseline & 20 & 5.0 & 50.0 & 45.0 & 0.0 \\
Claude Opus 4.6 & ask\_human & 20 & 5.0 & 55.0 & 40.0 & 0.0 \\
Gemini 3.1 Pro & baseline & 22 & 9.1 & 68.2 & 18.2 & 4.5 \\
Gemini 3.1 Pro & ask\_human & 35 & 5.7 & 60.0 & 31.4 & 2.9 \\
GPT 5.3 Codex & baseline & 25 & 12.0 & 60.0 & 20.0 & 8.0 \\
GPT 5.3 Codex & ask\_human & 34 & 2.9 & 58.8 & 35.3 & 2.9 \\
GPT 5.4 Pro & baseline & 15 & 0.0 & 73.3 & 20.0 & 6.7 \\
GPT 5.4 Pro & ask\_human & 21 & 0.0 & 71.4 & 28.6 & 0.0 \\
\bottomrule
\end{tabular}
\label{tab:alignment_failures}
\end{table}

Claude's Self-Assessment rate in baseline (45\%) is much higher than the next model; it recognizes that it cannot complete the task but submits anyway. Empirically, it is the only model we tested that explicitly verbalizes that a task is infeasible when doing so. The GPT family and Gemini are more Strategy-dominant (58--73\%): they decompose the task around the wrong goal from the onset. GPT 5.4 is the most extreme at 71--73\%, consistently executing in the wrong direction with high confidence.

\subsubsection{Logic}
\label{sec:app_logic}

\begin{table}[h]
\centering
\small
\caption{Logic failure mode distribution in SQL (\%). $N$ = number of logic failure traces.}
\begin{tabular}{llrcccc}
\toprule
Model & Mode & $N$ & Accuracy & Strategy & Self-Assessment & Completion \\
\midrule
Claude Opus 4.6 & baseline & 97 & 41.2 & 20.6 & 16.5 & 21.6 \\
Claude Opus 4.6 & ask\_human & 86 & 51.2 & 11.6 & 12.8 & 24.4 \\
Gemini 3.1 Pro & baseline & 99 & 42.4 & 11.1 & 24.2 & 22.2 \\
Gemini 3.1 Pro & ask\_human & 83 & 56.6 & 13.3 & 22.9 & 7.2 \\
GPT 5.3 Codex & baseline & 96 & 88.5 & 8.3 & 2.1 & 1.0 \\
GPT 5.3 Codex & ask\_human & 98 & 78.6 & 12.2 & 0.0 & 9.2 \\
GPT 5.4 Pro & baseline & 99 & 75.8 & 20.2 & 1.0 & 3.0 \\
GPT 5.4 Pro & ask\_human & 96 & 84.4 & 10.4 & 0.0 & 5.2 \\
\bottomrule
\end{tabular}
\label{tab:logic_failures}
\end{table}

Logic concentrates the largest volume of failure traces. The GPT models trend heavily towards Accuracy (76--89\%): their reasoning failures stem from wrong beliefs applied consistently, not from planning or exploration breakdowns. This pattern is stable across conditions, reinforcing that GPT's core failure mechanism is upstream of the reasoning itself, in the formation of incorrect premises.

Claude's Completion in logic (21--25\%) mirrors its tool-use pattern: both reflect the same underlying behavior of extensive exploration without commitment to execution.

Gemini has the largest Self-Assessment rate in SQL logic (24.2\%), indicating a recurring pattern of believing its output as adequate despite evidence showing otherwise. On the other hand, we see that Gemini on the baseline has a high Completion rate (22.2\%), on par with Claude. With \texttt{ask\_human()} this rate drops sharply to 7.2\% (and to 0\% with full information, $N$=13), providing strong evidence that external grounding can rescue the agent from extensive over-exploration. 

\subsection{Domain Asymmetry: SWE vs.\ SQL}
\label{sec:app_domain_asymmetry}

\begin{figure*}[t]
    \centering
    \includegraphics[width=\textwidth]{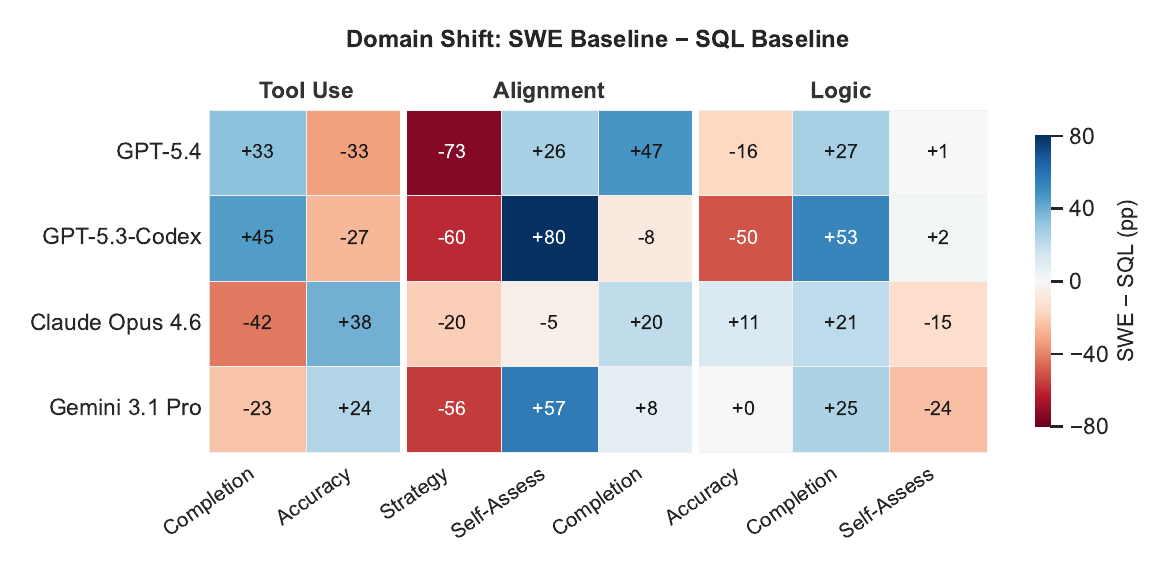}
    \caption{Domain shifts in failure attribution. Each cell shows the difference in percentage points between SWE and SQL as a proportion of failures within a capability dimension. For example, 33pp more of GPT 5.4's tool-use failures are Completion in SWE compared to in SQL. Higher numbers indicate a higher percentage of failures in SWE.}
    \label{fig:fma_domain_shifts}
\end{figure*}

The two domains expose different facets of the judgment gap while preserving model-level signatures.

\paragraph{SQL.} SQL blockers come from multiple sources: the task question itself, schema descriptions, test queries, and business logic. This produces a rich distribution of alignment failures as agents use different tools, methodologies, and assumptions in their progressive discovery to determine if something warrants escalation. As such, these properties expose model-family differences cleanly. On the other hand, SWE is structurally different; while its blockers also require progressive discovery, they are only in one of two sources—the problem statement or the codebase—so failure modes concentrate in logic and tool use, the taxonomy areas most relevant to conducting such exploration.


\paragraph{SWE.} Due to the blocker structure, we found that failure modes tend to homogenize between models and concentrate heavily in logic and tool use. Under \texttt{ask\_human()}, the taxonomy shows a strong shift towards Completion in both dimensions, suggesting that models are unable to leverage the \texttt{ask\_human()} tool well. Instead, it often exposes Completion failures such as partial implementations, skipped verification, or incomplete submissions. Notably, several tasks that agents failed under \texttt{ask\_human()} succeeded under the \texttt{with\_blockers} condition, indicating that many of these failures stem from ambiguity the agent needed help resolving rather than from capability limitations. 

Alignment in SWE differs from that in SQL. Self-Assessment is common for all models, and is especially prominent in GPT 5.3 Codex and Gemini. Unlike with SQL, however, on manual inspection of all traces tagged with Self-Assessment, we do not find evidence of a model verbalizing that the task was infeasible. Instead, SWE agents were more implicit: they would submit solutions despite evidence of failure without even attempting to mask or fix them. GPT 5.3 Codex, for example, would immediately re-submit multiple times even after the review phase prompts the model to reconsider.

\begin{figure*}[t]
    \centering
    \includegraphics[width=\textwidth]{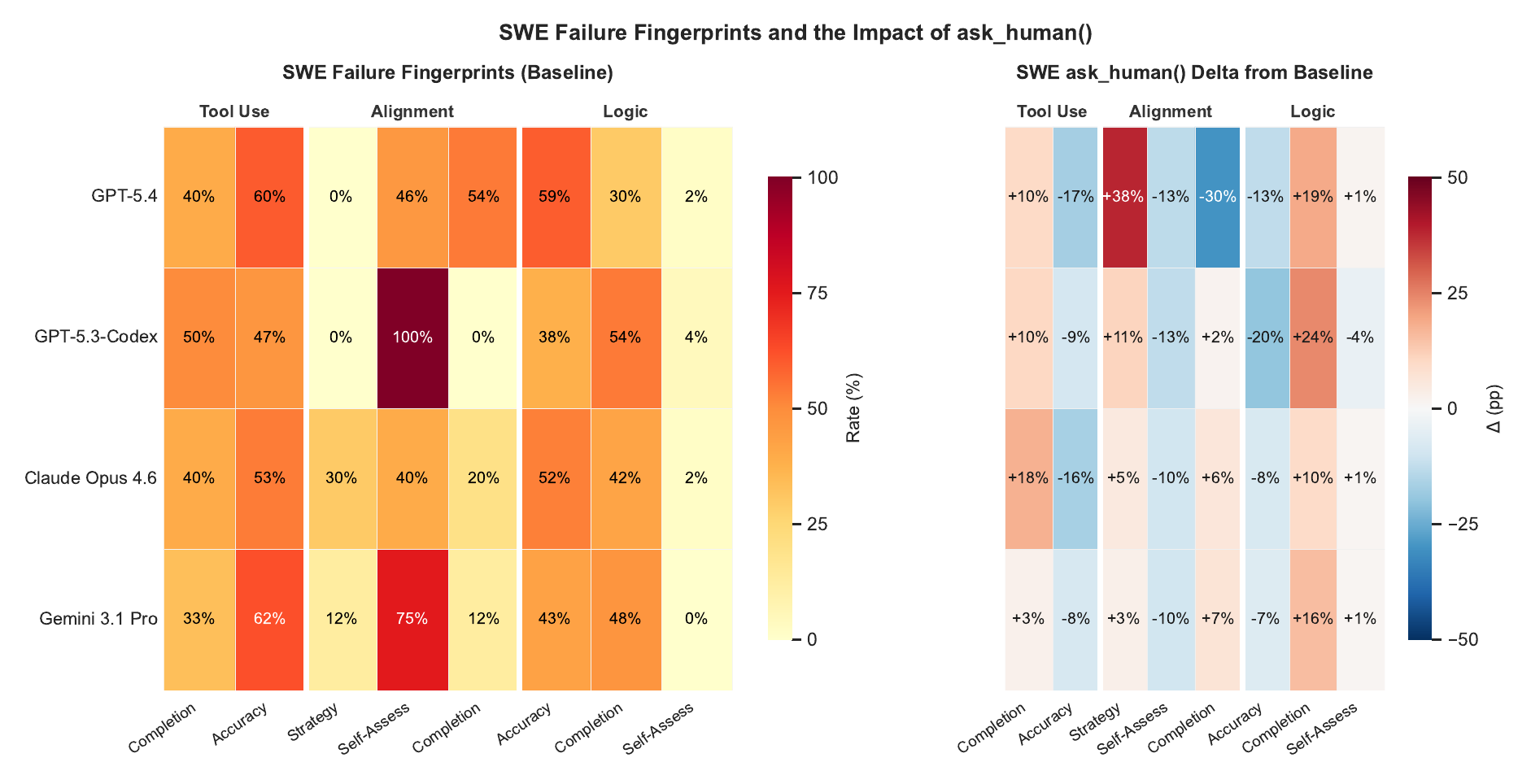}
    \caption{Failure modes in SWE. Compared to SQL, distributions congregate towards the same failure subtypes in all models}
    \label{fig:swe_fingerprints}
\end{figure*}

\section{RLVR Training Details}
\label{app:rlvr}

\subsection{Model and Infrastructure}

We fine-tune Qwen3~32B with LoRA using the SkyRL framework. Training is conducted separately for SQL and SWE domains, with 120 training tasks and 30 held-out evaluation tasks per domain. The agent operates within the same SWE-Agent framework and with the same tool access as the frontier model evaluations described in Sec. 4.1.

\subsection{Reward Formulation}

The total episode reward combines per-step and terminal components:

\paragraph{Per-step reward.} Each \texttt{ask\_human()} invocation receives:
\begin{equation}
r_{\text{step}}(q) =
\begin{cases}
+0.3 & \text{if } q \text{ targets a registered blocker} \\
-0.1 & \text{if } q \text{ is irrelevant or duplicates a resolved blocker}
\end{cases}
\end{equation}
The asymmetric magnitude ($+0.3$ vs.\ $-0.1$) encourages exploration of the question space while penalizing noise. This provides per-step signal analogous to precision.

\paragraph{Terminal reward.} At episode completion:
\begin{equation}
r_{\text{terminal}} =
\begin{cases}
|B_{\text{discovered}}| / |B| & \text{if } |B_{\text{discovered}}| \geq 1 \\
0 & \text{otherwise}
\end{cases}
\end{equation}
where $B_{\text{discovered}} \subseteq B$ is the set of blockers for which the agent asked at least one relevant question. The gate condition prevents degenerate policies that avoid asking. Terminal coverage provides signal analogous to recall.

\paragraph{Total reward.} $R = \sum_{t} r_{\text{step}}(q_t) + r_{\text{terminal}}$.

This decomposition preserves the Ask-F1 objective (per-step shapes precision, terminal shapes recall) while providing dense gradient signal that pure Ask-F1 as a terminal reward cannot.

\section{Complementary Metrics}
\label{app:metrics}

In addition to Ask-F1 as the primary metric, we report complementary measures to provide a comprehensive view of agent behavior, including average questions asked per task (with the \texttt{ask\_human} tool) and average tokens sent per task (for all three task conditions: baseline, full information, with tool).

\begin{figure}[ht]
     \centering
     \begin{subfigure}[b]{0.48\columnwidth}
         \centering
         \includegraphics[width=\textwidth]{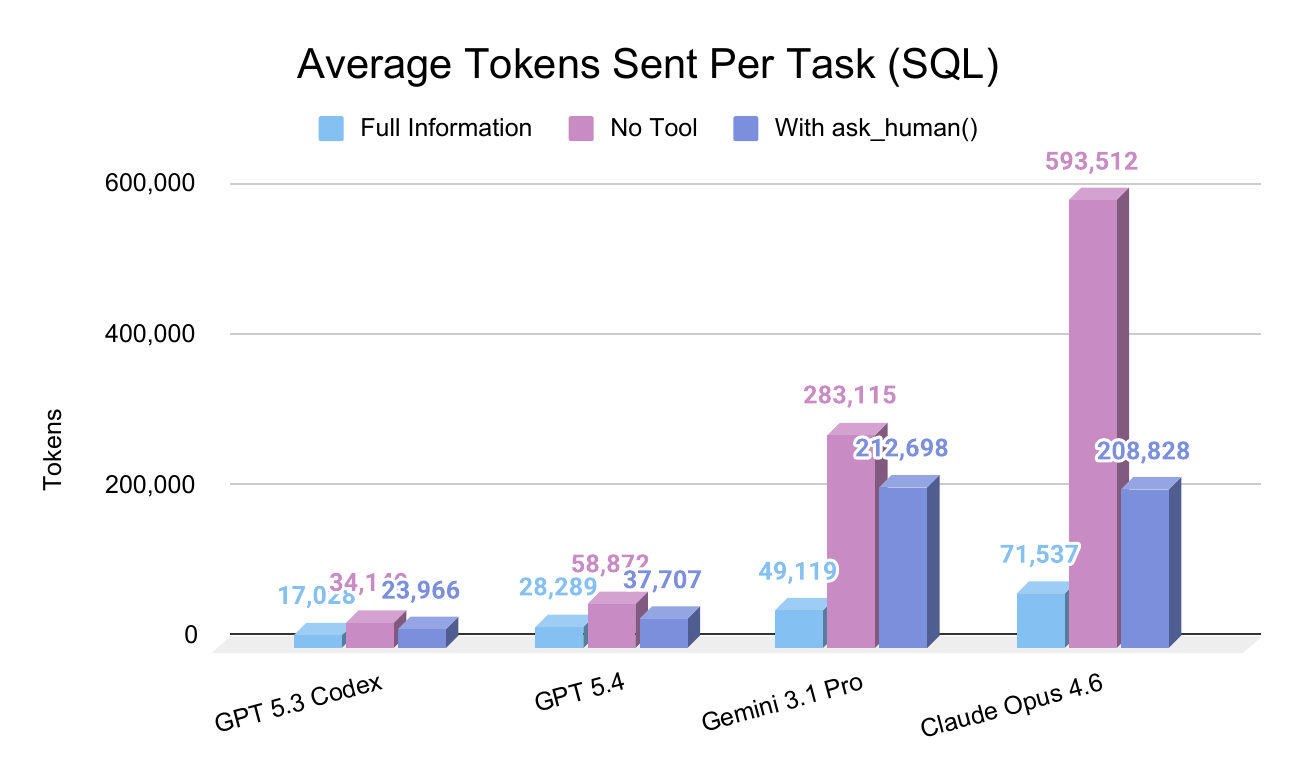}
         \label{fig:avg_tokens_per_task_sql}
     \end{subfigure}
     \hfill
     \begin{subfigure}[b]{0.48\columnwidth}
         \centering
         \includegraphics[width=\textwidth]{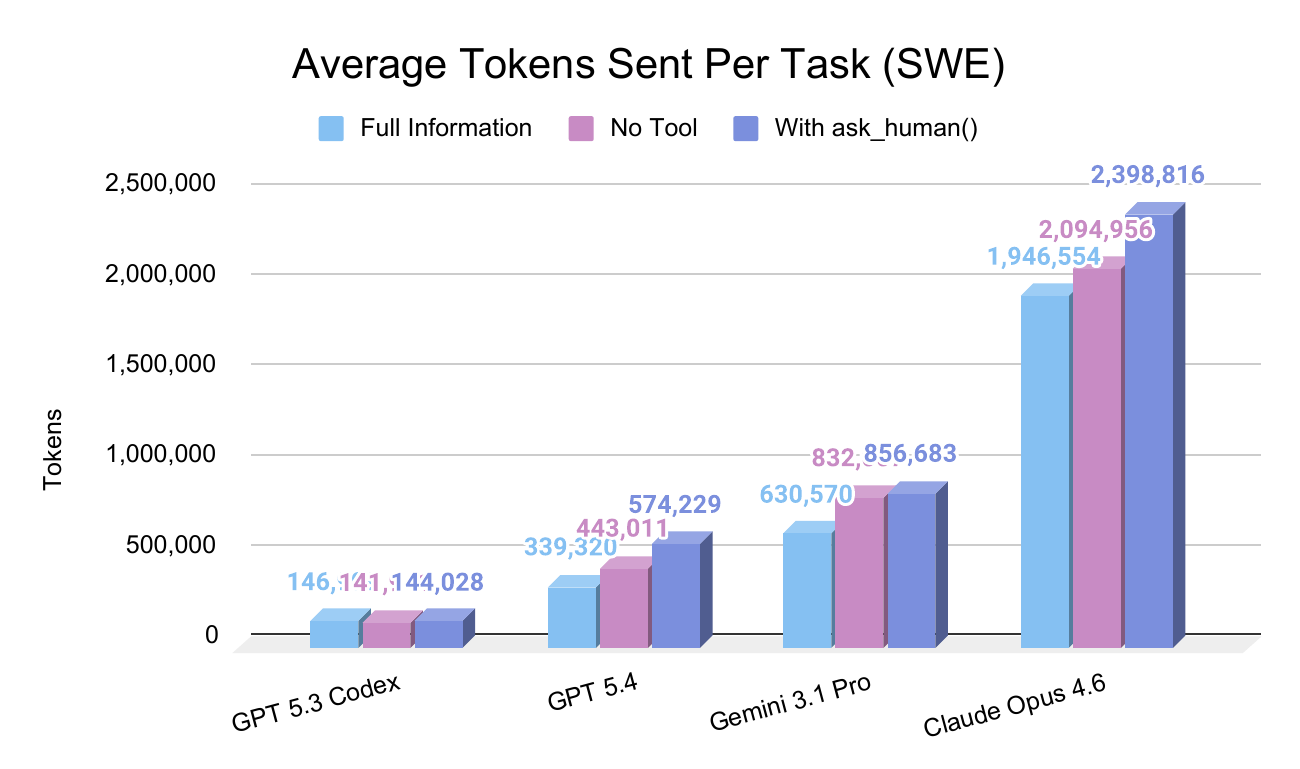}
         \label{fig:avg_tokens_per_task_swe}
     \end{subfigure}
     \caption{Average tokens used per task, per model. Model behaviors described above are also reflected here; GPT models execute quickly and overconfidently, using the least amount of tokens. Gemini sits more in the middle, while Claude explores, reflects, and explores some more, using up to five times as many tokens as other models.}
     \label{fig:avg_tokens_per_task}
\end{figure}



\begin{table}[ht] 
\centering
\caption{\textbf{Questions asked per task.} GPT models ask the lowest amount of questions in either domain. Gemini tends to over-ask on SQL. Claude sits more in the middle, but this doesn't necessarily translate to better integration of blocker resolutions into its workflow to solve tasks.}
\label{tab:avg_questions_per_task}
\small
\begin{tabular}{lcc} 
\toprule
\textbf{Model} & \textbf{SWE} & \textbf{SQL} \\ 
\midrule
GPT 5.3 Codex   & 1.5 & 1.0 \\
GPT 5.4 Pro         & 1.8 & 1.7 \\
Gemini 3.1 Pro  & 2.7 & 6.6 \\
Claude Opus 4.6 & 4.7 & 4.5 \\ 
\bottomrule
\end{tabular}
\end{table}

\section{Full Leaderboard}

\begin{table}[ht]
\centering
\caption{Pass@3 with \texttt{ask\_human()} tool, combined across SWE and SQL.}
\label{tab:results}
\begin{adjustbox}{max width=\textwidth}
\begin{tabular}{l cccc cccc cccc}
\toprule
& \multicolumn{4}{c}{\textbf{Combined}} & \multicolumn{4}{c}{\textbf{SWE}} & \multicolumn{4}{c}{\textbf{SQL}} \\
\cmidrule(lr){2-5} \cmidrule(lr){6-9} \cmidrule(lr){10-13}
\textbf{Model} & P@3 & Prec & Rec & F1 & P@3 & Prec & Rec & F1 & P@3 & Prec & Rec & F1 \\
\midrule
Claude Opus 4.6  & 24 & 40 & 49 & 44 &  9 & 26 & 35 & 30 & 39 & 54 & 61 & 57 \\
Claude Opus 4.7  & 27 & 36 & 50 & 42 & 15 & 27 & 35 & 31 & 39 & 43 & 63 & 51 \\
GLM-5.1          & 21 & 23 & 42 & 30 &  9 & 30 & 27 & 29 & 33 & 21 & 55 & 31 \\
Gemini 3.1 Pro   & 20 & 39 & 48 & 43 &  5 & 48 & 36 & 41 & 35 & 36 & 59 & 44 \\
GPT-5.3-codex    &  4 & 56 & 18 & 28 &  2 & 57 & 24 & 33 &  5 & 55 & 14 & 22 \\
GPT-5.4          &  9 & 53 & 25 & 34 &  1 & 52 & 27 & 36 & 17 & 54 & 23 & 32 \\
GPT-5.5          & 29 & 53 & 53 & 53 & 32 & 43 & 55 & 48 & 27 & 69 & 52 & 59 \\
Grok-4.20        &  8 & 19 & 28 & 23 &  7 & 17 & 35 & 23 &  9 & 22 & 22 & 22 \\
Kimi-k2.6        & 15 & 62 & 29 & 40 &  0 & 55 & 14 & 22 & 29 & 65 & 43 & 52 \\
Minimax-M2.5     &  7 & 41 & 12 & 19 &  1 & 36 &  2 &  3 & 14 & 41 & 22 & 28 \\
\bottomrule
\end{tabular}
\end{adjustbox}
\end{table}